%% file: main.tex
\documentclass[11pt]{article}

\usepackage[preprint]{acl}

\usepackage{times}
\usepackage{latexsym}
\usepackage[T1]{fontenc}
\usepackage[utf8]{inputenc}
\usepackage{microtype}
\usepackage{inconsolata}
\usepackage{graphicx}
\usepackage{booktabs}
\usepackage{multirow}
\usepackage{amsmath}
\usepackage{amssymb}
\usepackage{listings}
\definecolor{promptboxbg}{gray}{0.97}
\lstdefinestyle{promptstyle}{
  basicstyle=\tiny\ttfamily,
  breaklines=true,
  backgroundcolor=\color{promptboxbg},
  frame=single,
  framerule=0.5pt,
  framesep=5pt,
  tabsize=1,
  showstringspaces=false
}

\title{From Refuse to Richness: Rubric Rewards for Long-Form Hallucination Reinforcement Learning}

\author{
     Yudong Wang$^{\dagger\ddagger}$ ~~~~~Zhe Yang$^{\dagger}$ ~~~~~Wenhan Ma$^{\dagger\ddagger}$ ~~~~~Rang Li$^{\dagger\ddagger}$ ~~~~~Qibin Yang$^{\dagger\ddagger}$ \\
     ~~~~~\textbf{Weimin Xiong$^{\dagger\ddagger}$} ~~~~~\textbf{Jiangshan Duo$^{\dagger\ddagger}$} ~~~~~\textbf{Zhifang Sui$^{\dagger}$} ~~~~~\textbf{Liang Zhao$^{\ddagger}$} \\
     $^{\dagger}$State Key Laboratory of Multimedia Information Processing, \\ School of Computer Science, Peking University \\
     $^{\ddagger}$LLM-Core Xiaomi
}

\begin{document}
\maketitle

\begin{abstract}
Rewards that penalize unsupported claims can improve grounding in long-form generation, but they can also teach models to answer less. We study this refusal-to-richness trade-off in long-form hallucination RL. Instead of using global richness proxies such as length, claim count, detail, or pairwise relevance, we represent each question with a key-point rubric that specifies the required and optional information a useful answer should cover. These rubrics define coverage directly and are used both for evaluation and as reward signals. Across grounding-only, proxy-based, rubric-only, and combined rewards, we find a stable trade-off: strict grounding rewards improve support but suppress coverage, while unconstrained rubric rewards improve coverage but weaken grounding. A soft combination of grounding, rubric coverage, and relevance gives the best balance in our experiments, improving in-distribution support while transferring better to out-of-distribution checklist tasks than either grounding-only or rubric-only rewards.
\end{abstract}

\input{sections/01_introduction}
\input{sections/02_related_work}
\input{sections/04_rubric_pipeline}
\input{sections/05_rubric_validation}
\input{sections/06_rl_method}
\input{sections/07_experiments}
\input{sections/08_analysis}
\input{sections/10_conclusion}

\input{sections/09_limitations}
\section*{Ethical Considerations}
The data we utilized are open for research, and evaluated LLMs are all publicly available by either parameters or API calls. Therefore, we do not anticipate any ethical concerns in our research.

\bibliography{references}

\clearpage
\onecolumn
\appendix
\input{sections/appendix}

\end{document}

%% file: sections/01_introduction.tex
\section{Introduction}

\input{figures/fig1_pareto}

Hallucination in long-form generation has two faces: factual hallucination, where claims are inconsistent with world knowledge, and faithfulness hallucination, where claims are unsupported by a provided context \citep{ji2023survey,huang2025survey,bang2025hallulens}. Both are central to evaluating long-form question answering \citep{min2023factscore,wei2024long,jacovi2025facts}, and recent reasoning-trained models can hallucinate at higher rates than their non-reasoning counterparts despite better task accuracy \citep{yao2025reasoning,yang2025barrel}. In this paper, we address both modes simultaneously by training models with rewards that combine grounding\footnote{Throughout the paper, we use \emph{grounding} to mean that a claim is supported by the provided context, when one is available, or by reliable world knowledge otherwise; this unifies faithfulness and factuality under a single sentence-level signal. The \emph{fact-} prefix in our reward variant names (\textsc{FACT-ONLY}, \textsc{FACT-PROXY}, \textsc{FACT-RUBRIC}, etc.) is used in this same broad sense and refers to the same support signal as \emph{grounding}, not to world-knowledge factuality alone.} with question-specific information coverage. But grounding alone --- whether the claims a model makes are supported --- does not capture whether an answer is rich. A model can be grounded while omitting important information, and a model can appear detailed while adding redundant or tangential facts. Worse, rewards that overfit to grounding can damage out-of-distribution behavior on tasks that require checklist-style content completion rather than mere risk avoidance.

Existing richness signals are limited. Response length can be inflated by filler. Claim count can be inflated by trivial or irrelevant facts. Detail-level rewards, such as counting supported claims, do not distinguish central information from peripheral information. Pairwise relevance or helpfulness judgments \citep{chen2025learning} provide a holistic preference but do not identify which question-specific content is missing. Hard refusal-style rewards \citep{xu2024rejection,yang2024alignment} can suppress unsupported content by suppressing answers, trading grounding for usefulness --- in our experiments, FACT-ONLY RL training causes response length to drop sharply within a few RL steps, as models learn that the safest way to avoid unsupported claims is to say less.

Richness is question-specific. Different questions require different answer structures, key facts, and levels of detail, so a single global proxy cannot capture richness across long-form questions.

We use per-sample key-point rubrics to encode this information. Given a question and, when available, a reference or context, a rubric specifies the required and optional information that a rich answer should cover. A rubric is not a gold answer; it is a checklist for evaluating whether the response covers the question-specific content that matters. We construct rubrics for FACTS Grounding, LongFact, FActScore, CL-bench \citep{dou2026clbenchbenchmarkcontextlearning}, and a 4K RL training set, and use them both for evaluation and as RL reward signals.

Our experiments on Qwen3-4B and DeepSeek-R1-Distill-Llama-8B reveal a consistent in-distribution / out-of-distribution split. On grounding benchmarks, strict grounding and generic detail/relevance rewards improve support but encourage conservative or under-covered answers, while rubric rewards preserve more of the requested content. On out-of-distribution checklist and creative-writing prompts, this picture reverses: hard grounding gates suppress transfer, whereas RUBRIC-ONLY and soft FACT-RUBRIC rewards give the strongest completion and pairwise-preference results.

The composition of grounding and coverage matters. Multiplicatively gated rubric rewards inherit the out-of-distribution weakness of pure grounding rewards. Additive grounded-coverage rewards recover only part of the out-of-distribution gap. A soft three-component reward, combining a non-gating grounding term, rubric coverage, and pairwise relevance, gives the best balanced tradeoff in our current experiments. Grounding remains useful, but all-or-nothing gates can suppress the checklist-style behavior that long-form answers need.

Our contributions are:
\begin{itemize}
  \item We formalize richness for long-form answers as question-specific information coverage, and frame long-form hallucination RL as jointly mitigating factual and faithfulness hallucination.
  \item We build a multi-generator union pipeline for per-sample key-point rubrics, applied to four grounding benchmarks, CL-bench, and a 4K RL training set.
  \item Across Qwen3-4B \citep{yang2025qwen3} and DeepSeek-R1-Distill-Llama-8B \citep{guo2025deepseek}, soft fact--rubric compositions best balance in-distribution grounding and out-of-distribution checklist and creative-writing behavior.
\end{itemize}

%% file: figures/fig1_pareto.tex
\begin{figure}[t]
\centering
\includegraphics[width=\linewidth]{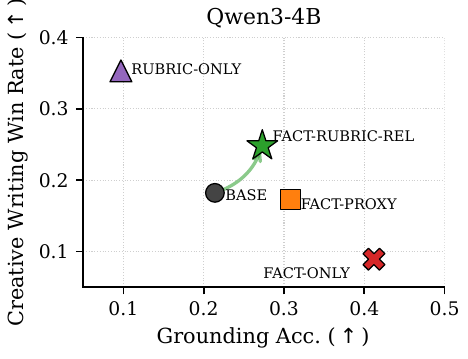}
\caption{In-distribution grounding accuracy versus out-of-distribution creative-writing win rate on Qwen3-4B; the green arrow marks the move from \textsc{BASE} to our \textsc{FACT-RUBRIC-REL} reward.}
\label{fig:pareto_qwen}
\end{figure}

%% file: sections/02_related_work.tex
\section{Related Work}

\paragraph{Long-form hallucination and grounding evaluation.}
Hallucination in language model output has been organized into taxonomies covering contextual unfaithfulness and factual error \citep{ji2023survey,huang2025survey,bang2025hallulens,li2023halueval}. Long-form factuality work operationalizes these failures by decomposing responses into claims and verifying them against evidence or reference documents, as in FActScore \citep{min2023factscore}, SAFE / LongFact \citep{wei2024long}, and FACTS Grounding \citep{jacovi2025facts}. Recent reasoning-style models can hallucinate more than their non-reasoning counterparts despite improved task accuracy \citep{yao2025reasoning,yang2025barrel,baker2025monitoring}, motivating careful reward design when applying RL to long-form generation. These protocols score what the model chooses to say, but not what a rich answer should contain. We add a question-specific coverage signal on top of grounding.

\paragraph{Reward design for grounded answers.}
A growing line of work uses RL or preference optimization to reduce hallucination and improve answer support \citep{lin2024flame,yang2024alignment,xu2024rejection,yin2023large,ren2025knowrl,wang2025enhancing}. \citet{song2025hallucination} document a ``hallucination tax'' from RL fine-tuning. Our closest baseline is \citet{chen2025learning} (LRF), which combines factual precision, supported-claim detail, and pairwise relevance against a reference response. We replace generic detail and relevance proxies with per-sample key-point rubrics, and study how composing grounding with coverage moves models along the in-distribution / out-of-distribution Pareto frontier.

\paragraph{Rubric-based evaluation.}
Rubrics and structured checklists have been used to evaluate open-ended generation and instruction following. Our focus is narrower: we build per-sample information-coverage rubrics for answering long-form questions, grounded in the question and the reference or context document when available, and use them as an RL reward.

%% file: sections/04_rubric_pipeline.tex
\section{Setup}

A rubric in this work is a checklist of question-specific information, not a reference answer: it specifies what content a rich response should cover for a given question, while leaving wording and order to the model. This yields a required-first coverage score $C_{\mathrm{rub}}(y, Q_x) \in [0,1]$ that complements the all-grounded indicator $G(y) \in \{0, 1\}$ used by grounding evaluation, with formal definitions deferred to Section~\ref{sec:reward}: $G$ measures whether all sentences are supported, while $C_{\mathrm{rub}}$ measures whether the answer covers what the question demands.

We instantiate the reward comparison on two open-weight model families, Qwen3-4B \citep{yang2025qwen3} and DeepSeek-R1-Distill-Llama-8B \citep{guo2025deepseek}. Within each family, the only experimental variable is the reward; all variants share the same base checkpoint, prompt pool, rollout configuration, evaluation splits, and judge models. We compare five reward variants summarized in Table~\ref{tab:reward_variants}, with formal definitions in Section~\ref{sec:reward}.

\begin{table*}[t]
\centering
\small
\begin{tabular}{lcccc}
\toprule
Method & Grounding & Detail & Relevance & Rubric \\
\midrule
FACT-ONLY & \checkmark &  &  &  \\
FACT-PROXY & \checkmark & \checkmark & \checkmark &  \\
FACT-RUBRIC & \checkmark &  &  & \checkmark \\
FACT-RUBRIC-REL & \checkmark &  & \checkmark & \checkmark \\
RUBRIC-ONLY &  &  &  & \checkmark \\
\bottomrule
\end{tabular}
\caption{Reward variants by active signal. The training setup is held fixed; only which signals enter the reward changes.}
\label{tab:reward_variants}
\end{table*}

\paragraph{Rubric construction.}
Each sample receives a structured rubric with a one-sentence completeness summary and a list of required and optional key points. Candidate rubrics for each prompt are generated independently by GPT-5.4 and Gemini-3-Flash-Preview, then merged by GPT-OSS-120B \citep{agarwal2025gptoss} into a single deduplicated union rubric. This uses different frontier model families as complementary rubric writers and keeps the final rubric broad enough to avoid single-generator omissions, while capping rubric size to keep coverage judging tractable. Required points represent core information that a rich answer should include; optional points represent useful but non-essential information.

\paragraph{Training and evaluation data.}
The RL training pool contains 4,000 prompts: 2,000 grounded prompts with context documents, sampled from the reliability-oriented training pool of \citet{wang2025enhancing}, and 2,000 open prompts without context, sampled from WildChat \citep{zhao2024wildchat} to broaden coverage beyond reference-grounded settings. We evaluate completed checkpoints on FACTS Grounding, LongFact, FActScore closed-book, and FActScore open-book as in-distribution grounding benchmarks. We also evaluate two out-of-distribution settings: (i) CL-bench \citep{dou2026clbenchbenchmarkcontextlearning}, a short-context benchmark scored by task-specific rubric checklists; and (ii) creative-writing prompts from Arena-Hard \citep{li2024crowdsourced,arenahard2024}, scored as pairwise preference against a fixed open-weight baseline by an LLM judge. Unless stated otherwise, in-distribution evaluation scores average Gemini-3.0-Flash-preview and GPT-OSS judgments \citep{agarwal2025gptoss}, then macro-average over benchmarks; CL-bench additionally averages a third GPT-5.1 judgment, and Arena-Hard creative writing uses GPT-4.1 as the pairwise judge. Implementation details and judge prompts are in Appendix~\ref{sec:appendix_rl_details}.

\paragraph{Metrics.}
We report three in-distribution metrics. \emph{Grounding accuracy} is $G(y) \in \{0, 1\}$, one if the answer has no unsupported or contradictory sentence and zero otherwise. \emph{Rubric coverage} is the required-first checklist score $C_{\mathrm{rub}}(y, Q_x)$ defined in Section~\ref{sec:reward}, gated by grounding for the main reported coverage figure. \emph{Claim number} is the average number of factual sentences judged as supported, unsupported, or contradictory. These metrics separately answer: is the response fully grounded, does it cover the prompt-specific information, and how much factual content does it attempt to provide.

Implementation details, exact hyperparameters, judge prompts, reward-router behavior, and checkpoint conversion are in Appendix~\ref{sec:appendix_rl_details}.

%% file: sections/05_rubric_validation.tex
\section{Reward Design}
\label{sec:reward}

We compare rewards that isolate grounding, generic richness proxies, rubric coverage, and their interaction (variants summarized in Table~\ref{tab:reward_variants}). All RL variants share the same model, prompt pool, format requirement, rollout setup, and judge family.

\paragraph{Grounding reward.}
Let $G(y)=1$ if no sentence in $y$ is judged unsupported or contradictory, and $G(y)=0$ otherwise. The FACT-ONLY reward is:
\begin{equation}
R_{\mathrm{fact}}(y) = G(y).
\end{equation}
This binary reward encourages the model to avoid unsupported content, but it does not reward covering more of the user's requested information.

\paragraph{Proxy baseline.}
We adapt the reward family from \citet{chen2025learning}. The original baseline combines factual precision, supported-claim detail, and pairwise preference against a reference answer. In our implementation, the support term is the same binary grounding indicator used by FACT-ONLY, so the baseline tests whether generic detail and relevance proxies improve beyond the grounding control:
\begin{equation}
R_{\mathrm{proxy}}(y) = G(y) + \lambda \log(1+F) + \mu R_{\mathrm{rel}},
\end{equation}
with $\lambda = \mu = 0.1$. Here $F$ is the number of supported sentences, and $R_{\mathrm{rel}}\in\{0,1\}$ is a pairwise win against a pre-generated answer from the base model. This baseline captures the intuition that richer answers should contain more supported content and be preferred holistically. Its limitation is that it still does not specify which content should be included.

\paragraph{Rubric coverage.}
For a rubric with $r_t$ required points and $o_t$ optional points, let $r_c$ and $o_c$ be the number of covered required and optional points. We use a gated coverage score:
\begin{equation}
C_{\mathrm{rub}} =
\begin{cases}
0.8\cdot \frac{r_c}{r_t}, & r_c < r_t, \\
0.8, & r_c=r_t,\ o_t=0, \\
0.8 + 0.2\cdot \frac{o_c}{o_t}, & r_c=r_t,\ o_t>0.
\end{cases}
\end{equation}
Required points therefore dominate the score, while optional points give additional credit only after the required checklist is complete.

\paragraph{Rubric rewards.}
The RUBRIC-ONLY ablation uses coverage without an explicit grounding term:
\begin{equation}
R_{\mathrm{rub}}(y)=C_{\mathrm{rub}}.
\end{equation}

We evaluate FACT-RUBRIC as an additive variant that combines grounding and rubric coverage without zeroing out rubric credit when grounding fails:
\begin{equation}
R_{\mathrm{fact+rub(add)}}(y)=
0.5G(y)+0.5C_{\mathrm{rub}}(y,Q_x).
\end{equation}
Finally, FACT-RUBRIC-REL adds a small relevance term:
\begin{equation}
\begin{split}
R_{\mathrm{fact+rub+rel}}(y)=
{}&0.5G(y)+0.4C_{\mathrm{rub}}(y,Q_x)\\
&+0.1R_{\mathrm{rel}}.
\end{split}
\end{equation}
These variants ask whether the out-of-distribution loss comes from using grounding information at all, or from composing grounding so strictly that completeness receives little direct credit. If RUBRIC-ONLY performs well, question-specific coverage alone carries useful behavior. If additive and relevance-augmented rubric rewards are stronger, then softer reward composition preserves more transferable coverage. Our experiments show both effects: on grounding-focused in-distribution evaluation, unconstrained rubric optimization hurts grounding; on CL-bench out-of-distribution evaluation, the softer rubric signal transfers better than hard grounding gates.

%% file: sections/06_rl_method.tex
\section{Training and Evaluation Protocol}
\label{sec:training}

We train each variant with GRPO \citep{shao2024deepseekmath}. Within each model family, all variants share the same base checkpoint, prompts, rollout configuration, and a hard response-format requirement: if the response omits a closing \texttt{</think>} tag or its post-reasoning answer is too short to support a meaningful grounding judgment, reward computation returns a fixed format penalty without calling the judge. The latter check matters empirically: in early DEEPSEEK-R1-Distill-Llama-8B FACT-ONLY runs we observed response length drop sharply within a few RL steps as the model learned that the safest way to avoid unsupported sentences is to say less. Full RL hyperparameters and infrastructure (verl-style HybridFlow \citep{sheng2024hybridflow}, vLLM rollouts \citep{kwon2023efficient}) are in Appendix~\ref{sec:appendix_rl_details}.

%% file: sections/07_experiments.tex
\section{Experiments}
\label{sec:experiments}

\begin{figure*}[!t]
\centering
\begin{minipage}{0.32\linewidth}
  \centering
  \includegraphics[width=\linewidth]{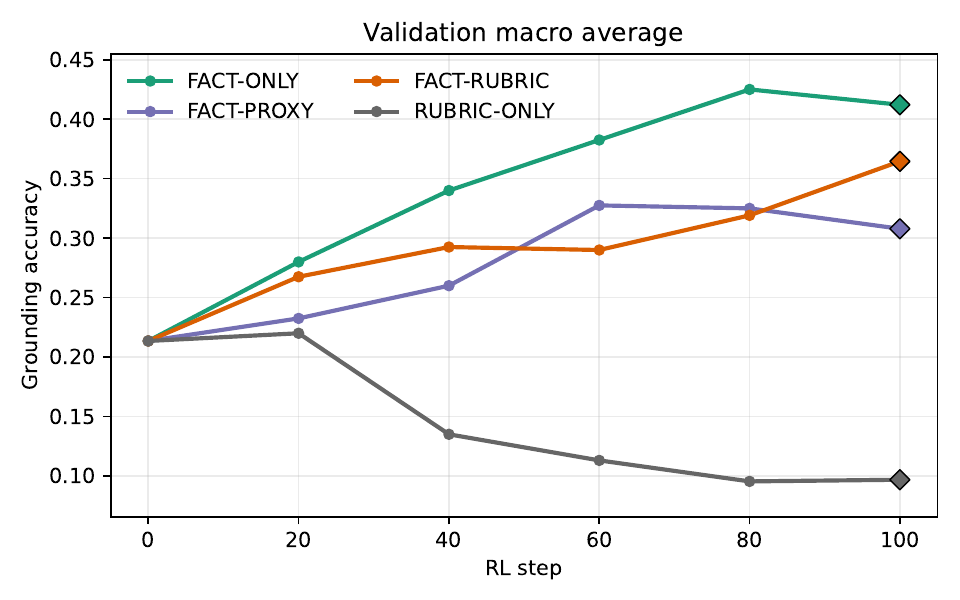}
\end{minipage}
\hfill
\begin{minipage}{0.32\linewidth}
  \centering
  \includegraphics[width=\linewidth]{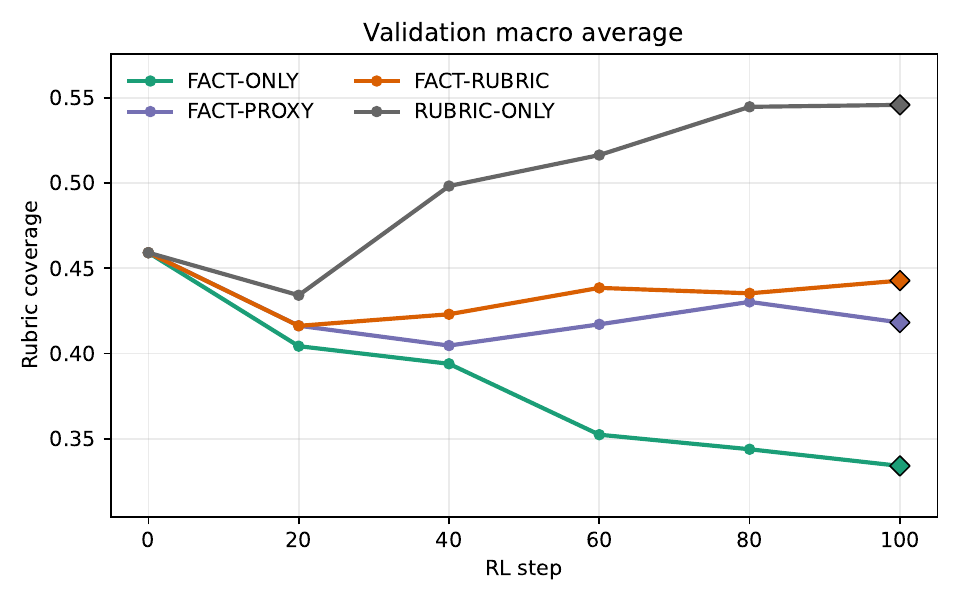}
\end{minipage}
\hfill
\begin{minipage}{0.32\linewidth}
  \centering
  \includegraphics[width=\linewidth]{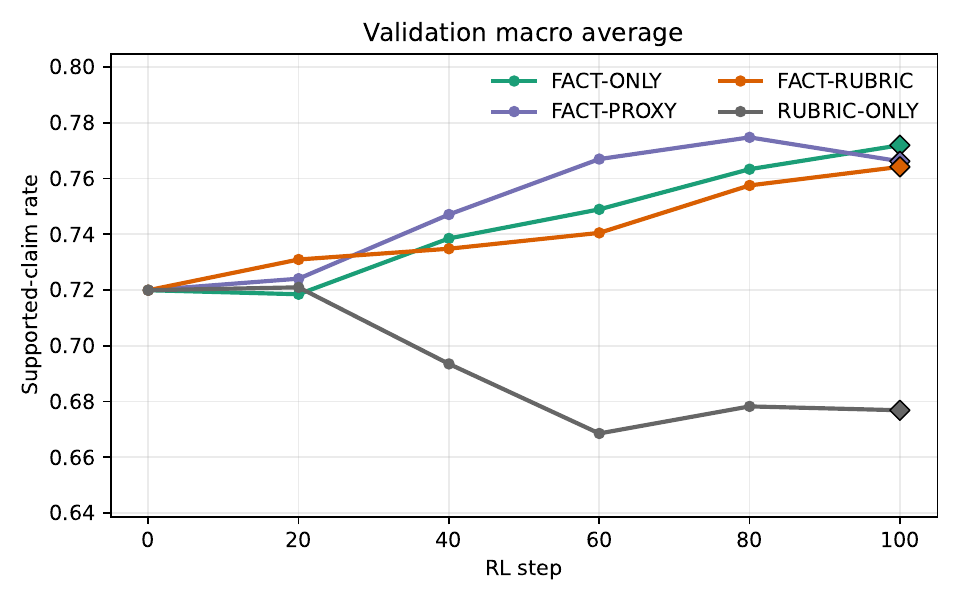}
\end{minipage}
\caption{Re-judged Qwen3-4B step curves for the originally-evaluated four reward variants (FACT-ONLY, FACT-PROXY, FACT-RUBRIC, RUBRIC-ONLY); FACT-RUBRIC-REL was added later and is reported only at its final checkpoint. Each point averages Gemini Flash and GPT-OSS judgments, macro-averaged over FACTS Grounding, LongFact, FActScore closed-book, and FActScore open-book.}
\label{fig:step_curves}
\end{figure*}

We evaluate the reward variants on Qwen3-4B and DeepSeek-R1-Distill-Llama-8B. The results combine training-time step curves for Qwen (Figure~\ref{fig:step_curves}), final held-out grounding evaluations, and out-of-distribution tests on CL-bench and Arena-Hard creative writing (Table~\ref{tab:combined_results}). Figure~\ref{fig:pareto_qwen} summarizes the same tension visually; here we separate the training dynamics from the final cross-model results.

\paragraph{Training dynamics.}
Figure~\ref{fig:step_curves} shows the four early-evaluated variants converging to clearly different operating points within the first hundred or so RL steps. FACT-ONLY ramps grounding accuracy fastest but rubric coverage stays at or below the base model. RUBRIC-ONLY moves in the opposite direction: rubric coverage rises sharply while grounding accuracy collapses. FACT-PROXY and FACT-RUBRIC track grounding more conservatively and keep rubric coverage closer to the base model. Claim support changes much less than the other two metrics across all variants, so the training curves mainly reflect changes in grounding and prompt-specific coverage.

\input{tables/combined_results}

\paragraph{In-distribution: grounding--coverage tradeoff.}
Table~\ref{tab:combined_results} confirms the expected grounding--coverage trade-off. On Qwen3-4B, FACT-ONLY reaches the highest grounding accuracy in the table but reduces both rubric coverage and claim number below the un-trained base. FACT-PROXY restores claim number but gives little coverage gain over the grounding control. RUBRIC-ONLY is the negative control: rubric coverage and claim number peak, but grounding collapses. The composed rewards sit between these extremes, and FACT-RUBRIC-REL keeps coverage and claim number high while still pulling grounding above base.

The DeepSeek-R1-Distill-Llama-8B replication (Table~\ref{tab:combined_results}, bottom) reproduces this ordering. FACT-ONLY and FACT-PROXY maximize grounding but drive rubric coverage noticeably below base. RUBRIC-ONLY again peaks coverage at the cost of grounding. FACT-RUBRIC-REL again sits balanced. The grounding--coverage trade-off is therefore a property of the reward composition, not of one base model.

\paragraph{Out-of-distribution: the picture reverses.}
The out-of-distribution columns reverse this ordering: the rewards that look safest on grounding benchmarks are the least transferable. On Qwen, FACT-ONLY reduces both creative-writing win rate and CL-bench solving rate below the base model, while RUBRIC-ONLY achieves the highest Qwen out-of-distribution numbers on both axes. The effect is sharper on DeepSeek: FACT-PROXY pushes grounding well above base but collapses creative writing and CL-bench to a small fraction of base-model performance. Hard grounding rewards therefore actively damage out-of-distribution behavior, rather than leaving it unchanged.

\paragraph{The Pareto frontier and FACT-RUBRIC-REL.}
The two views together describe the Pareto frontier shown in Figure~\ref{fig:pareto_qwen}. Strict grounding rewards sit at the high-grounding / low-OOD end; RUBRIC-ONLY sits at the opposite extreme. FACT-RUBRIC-REL is the only variant that improves over the base model on both in-distribution grounding and out-of-distribution behavior across both model families. It does not Pareto-dominate FACT-PROXY on in-distribution grounding alone, but the modest in-distribution drop is exchanged for substantial out-of-distribution gains, especially on DeepSeek. Rubric coverage thus drives out-of-distribution transfer, and a soft (non-gating) grounding term is needed to keep in-distribution grounding above the base model.

\paragraph{FActScore closed-book.}
Across all current variants, FActScore closed-book grounding accuracy stays near zero. We treat this split as orthogonal to the present comparison: closed-book performance is bottlenecked by the base model's parametric knowledge of the queried entities, which RL on this prompt pool does not address. Stronger entity knowledge, retrieval, or a closed-book-specific training and evaluation design would be required to move this number; we revisit this in Section~\ref{sec:limitations}.

%% file: tables/combined_results.tex
\begin{table*}[t]
\centering
\small
\setlength{\tabcolsep}{4pt}
\begin{tabular}{lrrrrr}
\toprule
Method & \multicolumn{3}{c}{In-distribution} & \multicolumn{2}{c}{Out-of-distribution} \\
\cmidrule(lr){2-4}\cmidrule(lr){5-6}
 & Grounding & Rubric Cov. & Claim Num. & CL-bench & Creative Writing \\
\midrule
\multicolumn{6}{l}{\textit{Qwen3-4B}} \\
BASE & 0.214 & 0.459 & 16.0 & 0.126 & 0.208 \\
FACT-ONLY & \textbf{0.412} & 0.334 & 11.7 & 0.090 & 0.108 \\
FACT-PROXY & 0.308 & 0.418 & 19.9 & 0.131 & 0.206 \\
FACT-RUBRIC & 0.258 & 0.480 & 23.0 & 0.135 & 0.180 \\
FACT-RUBRIC-REL & 0.273 & 0.499 & 24.9 & 0.136 & 0.292 \\
RUBRIC-ONLY & 0.097 & \textbf{0.546} & \textbf{36.0} & \textbf{0.159} & \textbf{0.376} \\
\addlinespace
\multicolumn{6}{l}{\textit{DeepSeek-R1-Distill-Llama-8B}} \\
BASE & 0.206 & 0.364 & 19.4 & 0.064 & 0.046 \\
FACT-ONLY & \textbf{0.512} & 0.163 & 8.6 & 0.017 & 0.016 \\
FACT-PROXY & 0.496 & 0.223 & 8.9 & 0.025 & 0.024 \\
FACT-RUBRIC & 0.499 & 0.457 & 17.3 & 0.075 & 0.028 \\
FACT-RUBRIC-REL & 0.382 & 0.461 & 17.8 & \textbf{0.095} & \textbf{0.058} \\
RUBRIC-ONLY & 0.149 & \textbf{0.528} & \textbf{34.5} & 0.092 & 0.056 \\
\bottomrule
\end{tabular}
\caption{Main results on in-distribution grounding benchmarks and out-of-distribution tasks.}
\label{tab:combined_results}
\end{table*}

%% file: sections/08_analysis.tex
\section{Analysis}

\paragraph{From refusal to richness.}
Strict grounding rewards create a simple incentive: unsupported content is penalized, but missing content is not. The model can improve the reward by answering only with claims it is confident are supported. FACT-ONLY makes this visible on Qwen3-4B, where grounding accuracy rises from 0.214 to 0.412 while rubric coverage drops from 0.459 to 0.334. On DPSK, the same reward also produces sharp response-length shrinkage in early RL steps. This is the refusal-to-richness problem in operational form: the reward must pay for what the answer should contain, not only for what it should avoid.

\paragraph{Why generic proxies do not close the gap.}
The LRF-style proxy reward adds two question-agnostic signals on top of grounding: $\log(1+F)$, which rewards additional supported sentences, and $R_{\mathrm{rel}}$, which rewards holistic preference over a base answer. On Qwen, FACT-PROXY raises grounding accuracy (0.308 vs.\ base 0.214), but rubric coverage remains close to base (0.418 vs.\ 0.459), and out-of-distribution behavior changes little. These proxies can reward grounded verbosity, but they do not identify the prompt-specific content that is missing.

\paragraph{What rubric coverage adds.}
Per-sample rubrics specify \emph{which} information a rich answer should cover. RUBRIC-ONLY gives the clearest isolation of this signal: it achieves the strongest out-of-distribution numbers while collapsing on in-distribution grounding. The model learns to satisfy explicit content requirements, which transfers to checklist tasks (CL-bench) and open-ended writing (Arena-Hard), but without a grounding term there is no pressure to keep every factual sentence supported.

\paragraph{Why soft composition retains both sides.}
Adding grounding back as a soft additive term avoids the all-or-nothing behavior of a multiplicative gate. Partially grounded but well-covered answers still receive some credit, so coverage is not erased by a single unsupported sentence. FACT-RUBRIC-REL combines $0.5\,G + 0.4\,C_{\mathrm{rub}} + 0.1\,R_{\mathrm{rel}}$; the small relevance term helps discourage high-coverage but low-quality answers. In our current runs, it is the only variant that improves over the base model on both in-distribution grounding and out-of-distribution behavior in both model families. It gives up some peak grounding relative to FACT-PROXY (Qwen 0.273 vs.\ 0.308), but keeps grounding above base while recovering much of RUBRIC-ONLY's out-of-distribution gain.

\paragraph{Cross-model consistency suggests a reward-level effect.}
The same ordering appears on Qwen3-4B and DeepSeek-R1-Distill-Llama-8B without retuning rewards or hyperparameters: RUBRIC-ONLY is strongest out-of-distribution, FACT-ONLY and FACT-PROXY are weakest there, and FACT-RUBRIC-REL is the only point that improves both sides over the base. The trade-off is therefore tied to reward composition rather than to one base checkpoint.

\paragraph{DPSK collapses harder than Qwen under hard grounding.}
Out-of-distribution damage from hard grounding rewards is more severe on DPSK than on Qwen. FACT-PROXY drops DPSK creative-writing win rate from 0.031 to 0.008 (roughly $4\times$) and CL-bench from 0.064 to 0.017, while the corresponding Qwen drops are about 50\%. Reasoning-distilled checkpoints already have a long-form generation style shaped by their distill objective; we conjecture that hard grounding rewards more readily over-write that style into a refusal-like mode, and that this effect compounds with the response-length collapse described below. A systematic study of why distillation-based reasoning models are more sensitive to hard grounding rewards is left to future work.

\paragraph{Length-collapse safeguard.}
Hard all-grounded rewards also create a simple reward-hacking path. In early DPSK FACT-ONLY runs, response length drops sharply within a few RL steps, as the model learns that shorter answers are less likely to contain unsupported sentences. We add a short-answer format penalty so the all-grounded condition cannot be satisfied by trivial truncation. This behavior is consistent with reports that RL fine-tuning can introduce a ``hallucination tax'' or other reward-hacking on reasoning-trained models \citep{song2025hallucination,baker2025monitoring,yao2025reasoning}.

%% file: sections/10_conclusion.tex
\section{Conclusion}

Hard grounding rewards make long-form answers safer but thinner; RUBRIC-ONLY rewards recover content coverage but lose grounding. Across Qwen3-4B and DeepSeek-R1-Distill-Llama-8B, the best trade-off comes from FACT-RUBRIC-REL, which combines grounding, rubric coverage, and relevance without an all-or-nothing gate. Per-sample rubrics are most useful as a complement to grounding, not as its replacement.

%% file: sections/09_limitations.tex
\section*{Limitations}
\label{sec:limitations}

The current cross-model check covers Qwen3-4B and DeepSeek-R1-Distill-Llama-8B, but both are relatively small open-weight models. Stronger proprietary or larger open-weight models may show different sensitivity to hard grounding gates.

%% file: sections/appendix.tex
\section*{Appendix}

\section{Use of LLM}

The authors only used the LLM to polish the language of this paper and help with coding.

\section{RL Implementation Details}
\label{sec:appendix_rl_details}

All online RL experiments use GRPO with the model's native chat template. Within each base-model family, all reward variants share the same training infrastructure, prompt pool, rollout configuration, optimizer settings, response-format requirement, and evaluation protocol. The only intended experimental difference is the reward computation.

\subsection{Rubric Pilot}

The rubric generator pilot is used only to select a practical construction recipe. On a 75-question pilot set, we found that rubrics generated by GPT-5.4 plus Gemini 3 Flash cover the answer content produced by a diverse set of models, including Claude Haiku, GPT, Gemini 3.1 Pro, GPT-OSS, and Claude Opus 4.6. We therefore use this two-generator recipe as a lightweight way to obtain broad per-question coverage before the merge and deduplication step.

\subsection{Rubric Construction Details}

Each rubric contains a one-sentence completeness summary and a list of key points. Each point has a short label, a required flag, a description, and context evidence when a reference is available:

\begin{quote}
\small
\texttt{\{id, point, required, description, context\_evidence\}}
\end{quote}

LongFact references are prefetched offline, so evaluation does not depend on live search. FActScore uses the full Wikipedia reference text because a small truncation audit found that 8K truncation often removes important biographical concepts. FACTS Grounding and CL-bench use the provided context documents directly.

\subsection{Prompt Templates}
\label{sec:appendix_prompts}

We use four main prompt families. The following are abridged versions of the actual templates, with task-specific fields shown in braces.

\paragraph{Rubric generation.}
This prompt asks a model to produce a per-question completeness checklist from the question and reference evidence.

\begin{lstlisting}[style=promptstyle]
Generate a completeness rubric for a
factual long-form question.

Input:
Question: {question}
Reference Document: {reference_document}

Instructions:
- Derive every point strictly from the reference.
- Each point must be specific, verifiable, and
  independently assessable.
- Include a short label, required flag, description,
  and context evidence for each point.
- Mark core information as required and useful
  non-essential information as optional.
- Aim for 8-20 points and never exceed 25.

Output JSON:
{
  "completeness_summary": "...",
  "rubric_points": [
    {
      "id": 1,
      "point": "...",
      "required": true,
      "description": "...",
      "context_evidence": "..."
    }
  ]
}
\end{lstlisting}

\paragraph{Rubric union.}
This prompt merges independently generated rubrics into one deduplicated union rubric.

\begin{lstlisting}[style=promptstyle]
Merge N candidate rubrics for the same question and
reference into one consolidated rubric.

Goals:
- Capture every distinct factual concept across the
  candidate rubrics.
- Deduplicate semantic equivalents.
- Preserve grounding evidence when a reference is
  available; use empty evidence for open prompts.
- Mark required if any source marks the concept
  required.
- Order by importance and cap the final rubric at
  25 points.

Output the same JSON schema as rubric generation.
\end{lstlisting}

\paragraph{Grounding and rubric-coverage judge.}
This is the main train-time and evaluation judge template. It returns sentence-level grounding labels and per-point coverage labels in one call.

\begin{lstlisting}[style=promptstyle]
You are given:
Context: {context_document}
User Query: {user_request}
Rubric: {rubric}
Response: {response}

Step 1: Split the response into sentences.
Label each sentence as:
- supported: directly entailed by the context
- unsupported: not entailed by the context
- contradictory: falsified by the context
- no_rad: does not require factual attribution

Be strict: if no clear excerpt supports a factual
sentence, label it unsupported.

Step 2: For each rubric point, label whether the
response covers it:
- covered
- missing

Output JSON:
{
  "sentences_check": [
    {
      "sentence": "...",
      "label": "<support_label>",
      "rationale": "...",
      "excerpt": "..." or null
    }
  ],
  "rubric_coverage": [
    {
      "rubric_id": 1,
      "point": "...",
      "required": true,
      "status": "covered|missing",
      "rationale": "..."
    }
  ],
  "overall_reasoning": "...",
  "has_formatting_errors": false,
  "request_completed": true
}
\end{lstlisting}

\paragraph{Pairwise relevance judge.}
This prompt is used for the relevance component in proxy and combined rewards.

\begin{lstlisting}[style=promptstyle]
You are an impartial judge evaluating two responses
to a user question.

Question: {question}
Response A: {response_a}
Response B: {response_b}

Choose the response that better follows the user's
instructions and provides a higher-quality answer.
Consider helpfulness, completeness, structure,
clarity, and instruction adherence.

Do not focus on factuality. Do not let length alone
decide. Treat ties as a win for Response A.

Output exactly one tag:
<answer>[[A]]</answer>
or
<answer>[[B]]</answer>
\end{lstlisting}

\subsection{Shared Hyperparameters}

\begin{table}[h]
\centering
\footnotesize
\begin{tabular}{ll}
\toprule
Setting & Value \\
\midrule
Base model & Qwen3-4B \\
Algorithm & GRPO \\
Nodes / GPUs & 4 nodes, 8 GPUs per node \\
Train batch size & 64 \\
Rollouts per prompt & 8 \\
Validation rollouts & 1 \\
Max prompt length & 32K tokens \\
Max response length & 16K tokens \\
Learning rate & $3\times 10^{-6}$ \\
Train temperature / top-$p$ & 1.0 / 1.0 \\
Validation temperature / top-$p$ & 0.6 / 0.95 \\
\bottomrule
\end{tabular}
\caption{Shared RL hyperparameters.}
\label{tab:appendix_rl_hparams}
\end{table}

The rollout backend uses vLLM with XFormers attention, chunked prefill enabled, tensor model parallel size 1, and GPU memory utilization 0.8. We enable FSDP parameter and optimizer offload, gradient checkpointing, Liger kernels, remove-padding, and dynamic batch sizing. Dynamic sampling is enabled with the \texttt{fine\_async\_reward} sampler, generation batch size 64, and hang timeout 900 seconds.

\subsection{Reward Router and Format Requirement}

The reward router dispatches each training sample to the reward function associated with its variant label. All train rewards impose the same hard format requirement: if a response does not contain a \texttt{<think>...</think>} block, the reward returns $-0.2$ without calling the judge.

Grounded rewards judge response sentences against the supplied context document and rubric. Open-mode rewards allow a world-knowledge fallback because the open prompts do not have a context document.

\input{tables/per_benchmark_results}
\subsection{Validation and Final Evaluation}

Online validation uses four fixed 50-sample subsets from FACTS Grounding, LongFact, FActScore closed-book, and FActScore open-book.
The validation scorer returns sentence-grounding accuracy as a scalar. For LongFact validation, \texttt{eval\_longfact\_50\_score} uses the lenient LongFact prompt with world-knowledge fallback, because cached LongFact references are not exhaustive.

The larger Setup-2 evaluation uses four 256-sample held-out subsets with the same benchmark split names. Final checkpoints are evaluated first, followed by periodic checkpoints for step-curve analysis.

%% file: tables/per_benchmark_results.tex
\subsection{Per-Benchmark Setup-2 Results}
\label{sec:appendix_per_benchmark}

Tables~\ref{tab:per_bench_facts_grounding}--\ref{tab:per_bench_factscore_open_book} break the macro-averaged Setup-2 numbers in Table~\ref{tab:combined_results} down by individual benchmark, and report the two judges (Gemini-3.0-Flash-preview and GPT-OSS-120B) separately rather than averaged. \emph{Claim Num.} is the average number of factual sentences (supported, unsupported, or contradictory) the judge identified per response, and serves as a length / verbosity proxy.

\begin{table*}[h]
\centering
\footnotesize
\begin{tabular}{llrrrrrr}
\toprule
 & & \multicolumn{2}{c}{Grounding Acc.} & \multicolumn{2}{c}{Rubric Cov.} & \multicolumn{2}{c}{Claim Num.} \\
\cmidrule(lr){3-4} \cmidrule(lr){5-6} \cmidrule(lr){7-8}
Model & Method & Gem. & GPT-OSS & Gem. & GPT-OSS & Gem. & GPT-OSS \\
\midrule
\multirow{6}{*}{Qwen3-4B}
& BASE            & 0.457 & 0.417 & 0.645 & 0.560 & 9.7  & 9.3  \\
& FACT-ONLY       & 0.855 & 0.770 & 0.456 & 0.401 & 4.6  & 4.6  \\
& FACT-PROXY      & 0.758 & 0.656 & 0.579 & 0.522 & 8.9  & 8.9  \\
& FACT-RUBRIC     & 0.637 & 0.582 & 0.750 & 0.699 & 14.0 & 13.5 \\
& FACT-RUBRIC-REL & 0.608 & 0.551 & 0.787 & 0.722 & 15.5 & 15.2 \\
& RUBRIC-ONLY     & 0.271 & 0.230 & 0.878 & 0.823 & 28.5 & 28.2 \\
\addlinespace
\multirow{6}{*}{DPSK-Llama-8B}
& BASE            & 0.398 & 0.336 & 0.652 & 0.591 & 16.4 & 20.3 \\
& FACT-ONLY       & 0.781 & 0.738 & 0.469 & 0.374 & 7.6  & 7.2  \\
& FACT-PROXY      & 0.655 & 0.637 & 0.365 & 0.282 & 7.9  & 7.1  \\
& FACT-RUBRIC     & 0.774 & 0.793 & 0.753 & 0.713 & 17.1 & 15.9 \\
& FACT-RUBRIC-REL & 0.773 & 0.761 & 0.738 & 0.657 & 13.7 & 13.6 \\
& RUBRIC-ONLY     & 0.193 & 0.385 & 0.883 & 0.833 & 27.3 & 37.1 \\
\bottomrule
\end{tabular}
\caption{Per-variant results on \textbf{FACTS Grounding}, reported separately for the two judges. Variants and steps follow Table~\ref{tab:combined_results}.}
\label{tab:per_bench_facts_grounding}
\end{table*}

\begin{table*}[h]
\centering
\footnotesize
\begin{tabular}{llrrrrrr}
\toprule
 & & \multicolumn{2}{c}{Grounding Acc.} & \multicolumn{2}{c}{Rubric Cov.} & \multicolumn{2}{c}{Claim Num.} \\
\cmidrule(lr){3-4} \cmidrule(lr){5-6} \cmidrule(lr){7-8}
Model & Method & Gem. & GPT-OSS & Gem. & GPT-OSS & Gem. & GPT-OSS \\
\midrule
\multirow{6}{*}{Qwen3-4B}
& BASE            & 0.149 & 0.196 & 0.268 & 0.241 & 23.9 & 22.3 \\
& FACT-ONLY       & 0.324 & 0.332 & 0.289 & 0.242 & 20.3 & 19.9 \\
& FACT-PROXY      & 0.238 & 0.254 & 0.330 & 0.265 & 28.5 & 27.6 \\
& FACT-RUBRIC     & 0.161 & 0.211 & 0.342 & 0.300 & 31.7 & 31.0 \\
& FACT-RUBRIC-REL & 0.191 & 0.246 & 0.353 & 0.304 & 33.3 & 32.4 \\
& RUBRIC-ONLY     & 0.082 & 0.125 & 0.421 & 0.354 & 46.1 & 47.2 \\
\addlinespace
\multirow{6}{*}{DPSK-Llama-8B}
& BASE            & 0.242 & 0.203 & 0.231 & 0.204 & 21.0 & 31.6 \\
& FACT-ONLY       & 0.473 & 0.410 & 0.139 & 0.128 & 14.8 & 16.0 \\
& FACT-PROXY      & 0.424 & 0.430 & 0.157 & 0.134 & 18.2 & 15.6 \\
& FACT-RUBRIC     & 0.391 & 0.375 & 0.191 & 0.172 & 15.9 & 18.3 \\
& FACT-RUBRIC-REL & 0.251 & 0.312 & 0.239 & 0.186 & 20.5 & 17.2 \\
& RUBRIC-ONLY     & 0.101 & 0.189 & 0.321 & 0.341 & 26.9 & 48.9 \\
\bottomrule
\end{tabular}
\caption{Per-variant results on \textbf{LongFact} (lenient world-knowledge-fallback prompt; see Section~\ref{sec:training}).}
\label{tab:per_bench_longfact}
\end{table*}

\begin{table*}[h]
\centering
\footnotesize
\begin{tabular}{llrrrrrr}
\toprule
 & & \multicolumn{2}{c}{Grounding Acc.} & \multicolumn{2}{c}{Rubric Cov.} & \multicolumn{2}{c}{Claim Num.} \\
\cmidrule(lr){3-4} \cmidrule(lr){5-6} \cmidrule(lr){7-8}
Model & Method & Gem. & GPT-OSS & Gem. & GPT-OSS & Gem. & GPT-OSS \\
\midrule
\multirow{6}{*}{Qwen3-4B}
& BASE            & 0.095 & 0.101 & 0.439 & 0.336 & 15.8 & 15.0 \\
& FACT-ONLY       & 0.004 & 0.008 & 0.066 & 0.007 & 8.6  & 7.5  \\
& FACT-PROXY      & 0.004 & 0.004 & 0.068 & 0.008 & 16.6 & 14.8 \\
& FACT-RUBRIC     & 0.000 & 0.000 & 0.064 & 0.011 & 19.6 & 17.4 \\
& FACT-RUBRIC-REL & 0.000 & 0.000 & 0.067 & 0.010 & 18.4 & 16.3 \\
& RUBRIC-ONLY     & 0.000 & 0.000 & 0.081 & 0.012 & 30.8 & 27.9 \\
\addlinespace
\multirow{6}{*}{DPSK-Llama-8B}
& BASE            & 0.000 & 0.000 & 0.094 & 0.011 & 13.9 & 13.0 \\
& FACT-ONLY       & 0.043 & 0.086 & 0.052 & 0.006 & 5.7  & 4.5  \\
& FACT-PROXY      & 0.156 & 0.414 & 0.046 & 0.005 & 4.6  & 2.6  \\
& FACT-RUBRIC     & 0.027 & 0.168 & 0.043 & 0.008 & 5.3  & 4.1  \\
& FACT-RUBRIC-REL & 0.035 & 0.168 & 0.106 & 0.012 & 10.9 & 8.9  \\
& RUBRIC-ONLY     & 0.004 & 0.012 & 0.094 & 0.013 & 26.6 & 35.5 \\
\bottomrule
\end{tabular}
\caption{Per-variant results on \textbf{FActScore (closed-book)}. Grounding accuracy stays close to zero across the board; we discuss this split as orthogonal to the present comparison in Section~\ref{sec:experiments}.}
\label{tab:per_bench_factscore_closed_book}
\end{table*}

\begin{table*}[h]
\centering
\footnotesize
\begin{tabular}{llrrrrrr}
\toprule
 & & \multicolumn{2}{c}{Grounding Acc.} & \multicolumn{2}{c}{Rubric Cov.} & \multicolumn{2}{c}{Claim Num.} \\
\cmidrule(lr){3-4} \cmidrule(lr){5-6} \cmidrule(lr){7-8}
Model & Method & Gem. & GPT-OSS & Gem. & GPT-OSS & Gem. & GPT-OSS \\
\midrule
\multirow{6}{*}{Qwen3-4B}
& BASE            & 0.174 & 0.119 & 0.824 & 0.360 & 16.7 & 15.0 \\
& FACT-ONLY       & 0.555 & 0.449 & 0.684 & 0.527 & 14.3 & 14.0 \\
& FACT-PROXY      & 0.252 & 0.297 & 0.844 & 0.730 & 27.8 & 26.3 \\
& FACT-RUBRIC     & 0.220 & 0.254 & 0.889 & 0.785 & 28.5 & 28.0 \\
& FACT-RUBRIC-REL & 0.207 & 0.379 & 0.918 & 0.831 & 34.4 & 33.4 \\
& RUBRIC-ONLY     & 0.005 & 0.061 & 0.941 & 0.857 & 39.5 & 39.9 \\
\addlinespace
\multirow{6}{*}{DPSK-Llama-8B}
& BASE            & 0.224 & 0.242 & 0.642 & 0.491 & 21.3 & 17.3 \\
& FACT-ONLY       & 0.754 & 0.680 & 0.410 & 0.207 & 10.2 & 5.5  \\
& FACT-PROXY      & 0.645 & 0.734 & 0.183 & 0.134 & 7.4  & 5.4  \\
& FACT-RUBRIC     & 0.719 & 0.749 & 0.916 & 0.858 & 28.2 & 33.6 \\
& FACT-RUBRIC-REL & 0.600 & 0.707 & 0.826 & 0.722 & 19.2 & 17.8 \\
& RUBRIC-ONLY     & 0.100 & 0.207 & 0.910 & 0.831 & 29.8 & 43.9 \\
\bottomrule
\end{tabular}
\caption{Per-variant results on \textbf{FActScore (open-book)}.}
\label{tab:per_bench_factscore_open_book}
\end{table*}